\documentclass{article}
\PassOptionsToPackage{numbers, compress}{natbib}
\usepackage[preprint]{neurips_2019}

\usepackage[utf8]{inputenc} 
\usepackage[T1]{fontenc}    
\usepackage{hyperref}       
\usepackage{url}            
\usepackage{booktabs}       
\usepackage{amsfonts}       
\usepackage{nicefrac}       
\usepackage{microtype}      

\usepackage{graphicx}
\usepackage{subfigure}
\usepackage{xcolor}
\usepackage{makecell}
\usepackage{amsmath}
\usepackage{multirow}
\usepackage[ruled]{algorithm2e}
\usepackage{xspace}
\usepackage{bm}

\usepackage[capitalise]{cleveref}

\usepackage{amsmath,amsfonts,bm}

\def\eqref#1{equation~\ref{#1}}

\def\1{\bm{1}}

\DeclareMathAlphabet{\mathsfit}{\encodingdefault}{\sfdefault}{m}{sl}
\SetMathAlphabet{\mathsfit}{bold}{\encodingdefault}{\sfdefault}{bx}{n}

\newcommand{\R}{\mathbb{R}}

\newcommand{\Cov}{\mathrm{Cov}}

\newcommand{\wavernn}{SparseWaveRNN\xspace}
\newcommand{\umnist}{MNIST-30k\xspace}
\newcommand{\bigmnist}{MNIST-500k\xspace}
\newcommand{\tfmnist}{MNIST-3M\xspace}

\newcommand{\cifar}{CIFAR-10\xspace}
\newcommand{\ciftinysep}{CIF-300k\xspace}
\newcommand{\cifsep}{CIF-900k\xspace}
\newcommand{\cifbig}{CIF-8M\xspace}

\newcommand{\wrnnsmall}{WR448\xspace}
\newcommand{\wrnnbig}{WR1792\xspace}

\newcommand{\snes}{SNES\xspace}
\newcommand{\es}{ES\xspace}
\newcommand{\cates}{C-ES\xspace}
\newcommand{\cmask}{FixMask\xspace}

\newcommand{\cateswr}{MS-wR\xspace}
\newcommand{\catesunif}{MS-wR+u\xspace}
\newcommand{\catesworbatched}{MS-woRb\xspace}
\newcommand{\catestopn}{MS-tN\xspace}

\newcommand{\esgst}{$n$\xspace}  
\newcommand{\esbst}{$b$\xspace}  
\newcommand{\esdimm}{d}  
\newcommand{\esgsm}{n}  
\newcommand{\esbsm}{b}  

\newcommand{\catnsm}{k}  

\newcommand*{\ie}{i.e.\@\xspace}

\renewcommand{\R}{\mathbb{R}} 
\newcommand{\mean}{\bm{\mu}} 
\renewcommand{\Cov}{\mathbf{\Sigma}} 
\newcommand{\stds}{\bm{\sigma}} 
\newcommand{\z}{\bm{z}} 

\newcommand{\logit}{\bm{l}}
\newcommand{\lval}[1]{l_{#1}}

\title{Non-Differentiable Supervised Learning with\\ Evolution Strategies and Hybrid Methods}

\author{%
  Karel Lenc\\
  DeepMind\\
  \texttt{lenck@google.com} \\
  \And
  Erich Elsen\\
  DeepMind\\
  \texttt{eriche@google.com} \\
  \And
  Tom Schaul\\
  DeepMind\\
  \texttt{schaul@google.com} \\
  \And
  Karen Simonyan\\
  DeepMind\\
  \texttt{simonyan@google.com} \\
}

\begin{document}

\maketitle

\begin{abstract}
In this work we show that Evolution Strategies (ES) are a viable method   
for learning non-differentiable parameters of large supervised models. ES are black-box optimization algorithms that estimate distributions of model parameters; however they have only been used for relatively small problems so far. We show that it is possible to scale ES to more complex tasks and models with millions of parameters.
While using ES for differentiable parameters is computationally impractical (although possible), we show that a hybrid approach is practically feasible in the case where the model has both differentiable and non-differentiable parameters. In this approach we use standard gradient-based methods for learning differentiable weights, while using ES for learning non-differentiable parameters -- in our case sparsity masks of the weights. This proposed method is surprisingly competitive, and when parallelized over multiple devices has only negligible training time overhead compared to training with gradient descent. Additionally, this method allows to train sparse models from the first training step, so they can be much larger than when using methods that require training dense models first.
We present results and analysis of supervised feed-forward models (such as MNIST and CIFAR-10 classification), as well as recurrent models, such as SparseWaveRNN for text-to-speech.
\end{abstract}

\section{Introduction}

Gradient-based optimization methods, such as SGD with momentum or Adam~\citep{kingma2014adam}, have become standard tools in the deep learning toolbox. While they are very effective for optimizing differentiable parameters, there is an interest in developing other efficient learning techniques that are complementary to gradient-based optimization. Evolution Strategies~\citep[ES,][]{Wierstra2014} is one such technique that has been used as an SGD alternative and has shown promising results on small-scale problems in reinforcement learning~\citep{igel2003neuroevolution, salimans2017evolution} and supervised learning~\citep{mandischer2002comparison, Lehman2017, Zhang2017, varelas2018comparative}.
ES is a black-box method and does not require the parameters to be differentiable. As such, it can potentially be applied to a much larger family of models than standard SGD.
Additionally, as ES only requires inference, it allows for training neural nets on inference-only hardware
(although we do not make use of this property here).
The goal of this paper is to explore how ES can be scaled up to larger and more complex models, from both an algorithmic and implementation perspective, and how it can be used in combination with SGD for training sparse neural networks.

We begin with investigating whether it is \emph{possible} to train \cifar classification ConvNets with ES (\cref{sec:snes}).
This is a harder task than training MNIST classification MLPs \citep[as in][]{Zhang2017}, and to address it with ES we develop a more efficient execution model which we call semi-updates. As a result, ES reaches competitive accuracy compared to SGD on \cifar, although this comes at the cost of significantly worse computational efficiency.
In~\cref{sec:hybrid} we then turn our attention to more practical applications (text-to-speech raw audio generation) and the use of ES alongside SGD for training large sparse models.  
Instead of relying on hand-crafted methods for training such models by pruning weights~\citep{NarangDSE17, Zhu2017},
we employ ES to learn the weight sparsity masks (i.e.\ which weight should be used in the final sparse model), while SGD is responsible for learning the values of the weights, and is performed in parallel with ES.
It turns out that unlike the ES for weights, the ES for sparsity patterns needs significantly fewer parameter samples which makes this approach computationally feasible.
Beyond the scientific interest in the feasibility of such hybrid learning techniques, one practical advantage of this approach is that it enables joint training of weights together with their sparsity mask, thus avoiding the need to first train a dense model (which might be too large to fit into memory), and then prune it.
The experiments on the state-of-the-art sparse text-to-speech model show that ES achieves comparable performance to SGD with pruning.

In summary, our contributions are three-fold: (i) we propose a new execution model of ES which allows us to scale it up to more complex ConvNets on \cifar;
(ii) we perform empirical analysis of the hyper-parameter selection for the ES training; (iii) we show how ES can be used in combination with SGD for hybrid training of sparse models, where the non-differentiable sparsity masks are learned by ES, and the differentiable weights by SGD.

\subsection*{Related work}
Prior works on hybrid SGD-ES methods mostly use ES for network structure and SGD to train the differentiable parameters of each sampled architecture \cite{yao1999evolving, igel2009genesis, real2018regularized}. This severely limits the size of the models where this method is practical as for each architecture sample the network has to be retrained.
Instead of meta-learning of the model, our goal is to optimize the model's non-differentiable and differentiable parameters \emph{jointly}. 
Recent works compare the relation between the SGD gradients and ES updates, either on synthetic landscapes \cite{Lehman2017} or MNIST models \cite{Zhang2017}, while \cite{maheswaranathan2018guided} uses SGD gradients to guide the sampler's search direction. Instead of combining them, we use SGD and ES for two different training tasks - SGD for the continuous, differentiable parameters and ES for modelling the sparsity mask distribution.

Sparse models, where a subset of the model parameters are set to exactly zero, offer a faster execution and compact storage in sparse-matrix formats. This typically reduces the model performance but allows deployment to resource-constrained environments \cite{kalchbrenner2018efficient, Theis2018FasterGP}.
Training unstructured sparse supervised models can be traced back at least to \cite{Cun90optimalbrain}, which used second order information to prune weights.
More tractable efforts can be traced at least to \cite{Strom97sparseconnection}, which pruned weights based on the magnitude and then retrained.  That was more recently extended in \citet{HanLMPPHD16} where the pruning and retraining process was carried out in multiple rounds. The pruning process was further refined in \cite{NarangDSE17} where the pruning gradually occurred during the first pass of training so that it took no longer to find a pruned model than to train the original dense model.  Finally, the number of hyper-parameters involved in the procedure was reduced in \cite{Zhu2017} 
and \citet{bellec2017deep} uses random walk for obtaining the sparsity mask. Additionally a number of Bayesian approaches have been proposed, such as Variational Dropout (VD) \cite{pmlr-v70-molchanov17a} or L0 regularization \cite{louizos2018learning}. VD uses reparametrization trick, while we learn directly the multinomial distribution with ES.
As indicated in \citet{gale19state}, VD and L0 regularization tends to perform not better than \cite{NarangDSE17} we compare against.

\section{Preliminaries}
\subsection{Natural Evolution Strategies}\label{ss:nes}

Black-box optimization methods are characterized by treating the objective function $f(\theta)$ as a black box, and thus do not require it to be differentiable with respect to its parameters $\theta$. Broadly speaking, they fall into three categories: population-based methods like genetic algorithms that maintain and adapt a set of parameters, distribution-based methods that maintain and adapt a distribution over the parameters $\theta \sim \pi$,
and Bayesian optimization methods that model the function $f$, using a Gaussian Process for example.

\emph{Natural evolution strategies} \citep[NES,][]{Wierstra2008,Wierstra2014} are an instance of the second type. NES proceeds by sampling a batch of $\esgst$ parameter vectors $\{\theta_i \sim \pi; 1 \leq i \leq \esgsm \}$, evaluating the objective for each ($f_i = f(\theta_i)$) and then updating the parameters of the distribution $\pi$ to maximize the objective $J = \mathbb{E}_{\theta \sim \pi}[f(\theta)]$, that is, the expectation of $f$ under $\pi$.
The attribute `natural' refers to the fact that NES follows the \emph{natural gradient} w.r.t. $J$ instead of the steepest (vanilla) gradient.

In its most common instantiation, NES is used to optimize unconstrained continuous parameters $\theta \in \R^d$, and employs a Gaussian distribution $\pi = \mathcal{N}(\mean, \Cov)$ with mean $\mean \in \R^d$ and covariance matrix $\Cov \in \R^{d \times d}$. 
For large $d$, estimating, updating and sampling from the full covariance matrix is too expensive, so here we use its diagonal approximation $\Cov = \operatorname{diag}(\stds^2)$ with element-wise variance terms $\stds^2 \in \R^d$; this variant is called Separable NES \citep[SNES,][]{snes}, and similar to \citep{ros2008simple}.
The sampled parameters are constructed as $\theta_i = \mean + \stds \odot \z_i$. where  $\odot$ is element-wise (Hadamard) product, from standard normal samples $\z \sim \mathcal{N}(0,\mathbf{I})$.
The natural gradient with respect to $(\mean, \stds)$ is given by:
\begin{eqnarray}\label{eq:snes}
\nabla_{\mean} J = \sum_{1 \leq i \leq n} u_i \z_i \notag, ~~
\nabla_{\stds} J = \stds \odot \sum_{1 \leq i \leq n} u_i \left(\z_i - 1 \right)^2
\end{eqnarray}
and the updates are
\begin{eqnarray}\label{eq:snes-update}
\mean \leftarrow \mean + \eta_{\mean} \stds \odot \nabla_{\mean} J & & \stds \leftarrow \stds \odot \exp(\frac{1}{2} \eta_{\stds} \nabla_{\stds} J )
\end{eqnarray}
where $\eta_{\mean}$ and $\eta_{\stds}$ are learning rates of the distribution's mean and variance parameters respectively, and $u_i$ is a transformation of the fitness $f_i$ (fitness-shaping, \cref{ss:fitness-shaping}).
Note that the multiplicative update on $\stds$ is guaranteed to preserve positive variances.

The more restricted case of constant variance ($\eta_{\stds}=0$), where the natural gradient coincides with the vanilla gradient, was advocated by~\citet{salimans2017evolution}, and it is sometimes referred to as simply `Evolution Strategies' (ES), even though it is subtly different from classic  Evolution Strategies~\citep{rechenberg1973evolutionsstrategie}: it matches the classic non-elitist $(1,\esgsm)$-ES only if the fitness shaping function is a top-$1$ threshold function.

\subsection{Fitness shaping functions}\label{ss:fitness-shaping}

It is often desirable to make evolution strategies more robust and scale-invariant by transforming the raw evaluation $f_i$ into a normalized utility $u_i$; this mapping is called \emph{fitness shaping}~\citep{hansen2001completely,Wierstra2014}. 
A common one is based on the rank of $f_i$ within the current batch, where $\nu$ is a hyper-parameter:
\begin{equation}\label{eq:fshaping}
    u_i = \frac{\max\left(0, \log\left(\frac{\esgsm}{\nu} + 1\right) - \log(\operatorname{rank}(f_i))\right)}
    {\sum_{j=1}^\esgsm \max\left(0, \log\left(\frac{\esgsm}{\nu} + 1\right) - \log(j)\right)} - \frac{1}{\esgsm}
\end{equation}
For $\nu = 2$, this method sets a constant utility to the lowest $50\%$ of the fitnesses, which we use in our experiments.

\section{Scaling up \snes for supervised models}\label{sec:snes}

In this section we propose a novel method of distributing the evaluation of \snes updates. In \cref{s:semi-updates} we present the semi-updates method, which allows for better distribution of random number generation, fitness evaluation, and update computation among multiple workers. This allows us to train large supervised models with millions of parameters without creating a bottleneck on the master that holds the distribution parameters. Further on, in \cref{ss:es-experiments} we investigate various hyper-parameter selections for obtaining competitive results.

\subsection{Speeding up \snes with semi-updates}\label{s:semi-updates}
In practice, \snes training of a model with $d$ parameters requires a large amount of weight samples per generation (generation size $\esgsm$) that grows with the dimension of its parameters $\theta \in \R^{\esdimm}$.
A \emph{standard} execution model for \snes is to draw the parameter samples $\{\theta_{1 \dots \esgsm}\}$ at the master process, which looks after distribution parameters $(\mean, \stds)$, and to distribute them among $\esbsm$ worker processes. Even with smaller models, the main bottleneck is generating $\esdimm \cdot \esgsm$ random values which are needed for computing the weighted sums (\cref{eq:snes}).

One possible improvement is to send only the initial distribution parameters $(\mean, \stds)$ and a random seed to the \esbst workers to let each worker generate $\esgsm / \esbsm$ parameter samples and exchange set of $\esgsm / \esbsm$ fitness scalars, similar to \cite{salimans2017evolution}; we refer to this method as \emph{batched} execution. This significantly reduces the amount of data communicated between the workers, however to perform an update, a worker which performs the ES update still needs to generate \emph{all} \esgst random parameter samples, which is slow for large generations and models.

We propose to use \emph{semi-updates} execution model. It is similar to the batched method as each worker obtains only the distribution parameters $(\mean, \stds)$, however instead of sending back the fitness scalars for each sample, it computes the update on the batch of  $\esgsm / \esbsm$ parameter samples according to Equations~\ref{eq:snes} and~\ref{eq:snes-update}, and sends it back to the master. 
Even though the standard ES execution model performs fitness shaping based on the rank of \emph{all} the parameter samples of a generation, 
doing it on only $\esgsm / \esbsm$ parameter samples within each worker has a surprisingly little effect on the final performance. The main advantage of this method is that each worker now has to generate only $\esgsm / \esbsm$ parameter samples while the master  performs only a simple average over $\esbsm$ semi-updates. However, contrary to the batched method, the new distribution parameters has to be communicated to each worker. 

\subsection{Experiments}\label{ss:es-experiments}
In this section we perform experiments with different variants of the \snes algorithm. In \cref{ss:semi-upd-exp}, we investigate the processing speed of different execution models defined in \cref{s:semi-updates}.
Later on, in \cref{ss:snes-cifar-exp} we investigate what is needed for being able to learn ConvNet models for \cifar classification with \snes. Finally, in \cref{ss:es-nsamples}, we investigate the dependence of accuracy on the generation size.

We use \snes for supervised classification models and the objective function $f$ is the mini-batch log-likelihood\footnote{It is possible to use accuracy, however it under-performs due to quantization of the mini-batch accuracy.}.
For all experiments we optimize all model parameters $\theta \in \R^\esdimm$ with a single normal distribution. The distribution parameters are updated with learning rates $\eta_{\mean}=1$ and $\eta_{\stds}= (3 + \ln{\esdimm}) / (5 \sqrt{\esdimm})$ \citep[p.~33]{Wierstra2014}. 
The mean $\mean$ is initialized using the truncated normal initializer \cite{glorot2010understanding}, and the variances $\stds^2$ are initialized to $1$.

We perform experiments on MNIST~\citep{lecun-98} and \cifar~\citep{krizhevsky2009learning} datasets using \mbox{ConvNet} models, which are described in \cref{ss:models}. Selected models differ mainly in the number of parameters, which is controlled by the number of hidden representations. For \cifar models, we train on random $24 \times 24$ crops and flips from the $32 \times 32$ training images. At test time, we process the full $32 \times 32$ test images and average-pool the activations before the classification layer. To reduce the parameter count, we use the Separable Convolution~\citep{chollet16xception}.

\subsection{Selected MNIST and CIFAR models}\label{ss:models}
In table \cref{tab:mnist_models} we provide details of the investigated MNIST and CIFAR models.
For the MNIST models, $\mathrm{C}_{\times n}^{M}$ stands for $M$ convolution filters of size $n \times n$, $\mathrm{P}$ for max pooling, and $\mathrm{F}_{a}^{b}$ for a fully connected layer with $b$ input features and $a$ output features. The \umnist model is identical to the model by \citet{Zhang2017}.

Each \cifar model consists of a set of $3\times3$ separable or standard convolutions where the number of filters per layer is specified in the column `Layers'. Layers with stride 2 are denoted as $s2$. The convolutional layers are followed by global average pooling and a fully connected layer with 10 outputs. The column `Sep' specifies whether separable or dense convolutions are used.

\begin{table}[h!]
    \centering
    \scriptsize
    \caption{MNIST models (left) and \cifar models (right) used in the experiments. ACC @10k and @80k denotes the SGD performance on the test set after 10k or 80k steps of training respectively.
    }
    \label{tab:mnist_models}
    \setlength{\tabcolsep}{1pt}
    \begin{minipage}[t]{.45\linewidth}    
        \begin{tabular}{l | r l | c } \toprule
        Name & \# Params & Layout & ACC @10k \\ \midrule
        \umnist & $28\,938$ & \makecell{
            $\mathrm{C}_{\times5}^{16}$, 
            $\mathrm{P}_{\times2}$,
            $\mathrm{C}_{\times2}^{32}$,\\
            $\mathrm{P}_{\times2}$,
            $\mathrm{F}_{10}^{1568}$}
        & $99.12\%$\\ \midrule
        
        \bigmnist & $454\,922$ & \makecell{
            $\mathrm{C}_{\times5}^{32}$, 
            $\mathrm{P}_{\times2}$,
            $\mathrm{C}_{\times2}^{64}$, \\
            $\mathrm{P}_{\times2}$,
            $\mathrm{F}_{128}^{3136}$,
            $\mathrm{F}_{10}^{128}$}
        & $99.28\%$\\ \midrule
        
        \tfmnist & $3\,274\,634$ & \makecell{
            $\mathrm{C}_{\times5}^{32}$, 
            $\mathrm{P}_{\times2}$,
            $\mathrm{C}_{\times2}^{64}$, \\
            $\mathrm{P}_{\times2}$,
            $\mathrm{F}_{1024}^{3136}$,
            $\mathrm{F}_{10}^{1024}$}
        & $99.33\%$\\
        \bottomrule
        \end{tabular}
    \end{minipage}
    \begin{minipage}[t]{.45\linewidth}
        \begin{tabular}{l  c c | c | r r } \toprule
        Name & Sep. & Layers & Params & \makecell{ACC\\@10k} & \makecell{ACC\\ @80k} \\ \midrule
        \ciftinysep & YES 
        & \makecell{
        $64$, $64$, \\
        $128_{s2}$, $128$, $128$,\\
        $256_{s2}$, $256$, $256$, \\
        $512_{s2}$} 
        & $358\,629$
        & $87.7$ & $91.7$\\ \midrule
    
        \cifsep  & YES 
        & \makecell{
        \ciftinysep \\ 
        + $512$, $512$ }
        & $895\,973$
        & $88.1$ & $92.01$\\ \midrule
        
        \cifbig  & NO & \makecell{
        \ciftinysep \\ + $512$, $512$}
        & $7\,790\,794$
        & $84.4$ & $94.6$\\
        
        \bottomrule
        \end{tabular}
    \end{minipage}
\end{table}

\subsubsection{Semi-updates}\label{ss:semi-upd-exp}
\begin{figure}
    \centering
    \setlength{\tabcolsep}{1pt}
    \begin{minipage}[t][][t]{.33\linewidth}
        \includegraphics[width=1\linewidth]{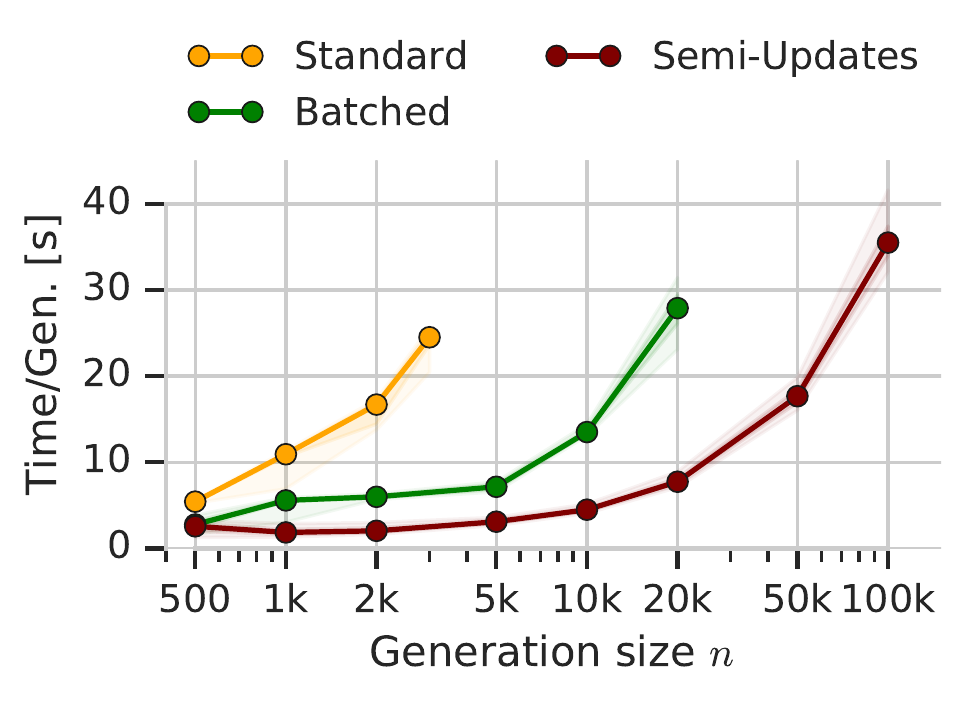}
    \end{minipage}%
    \begin{minipage}[t][][t]{.33\linewidth}
        \includegraphics[width=1\linewidth]{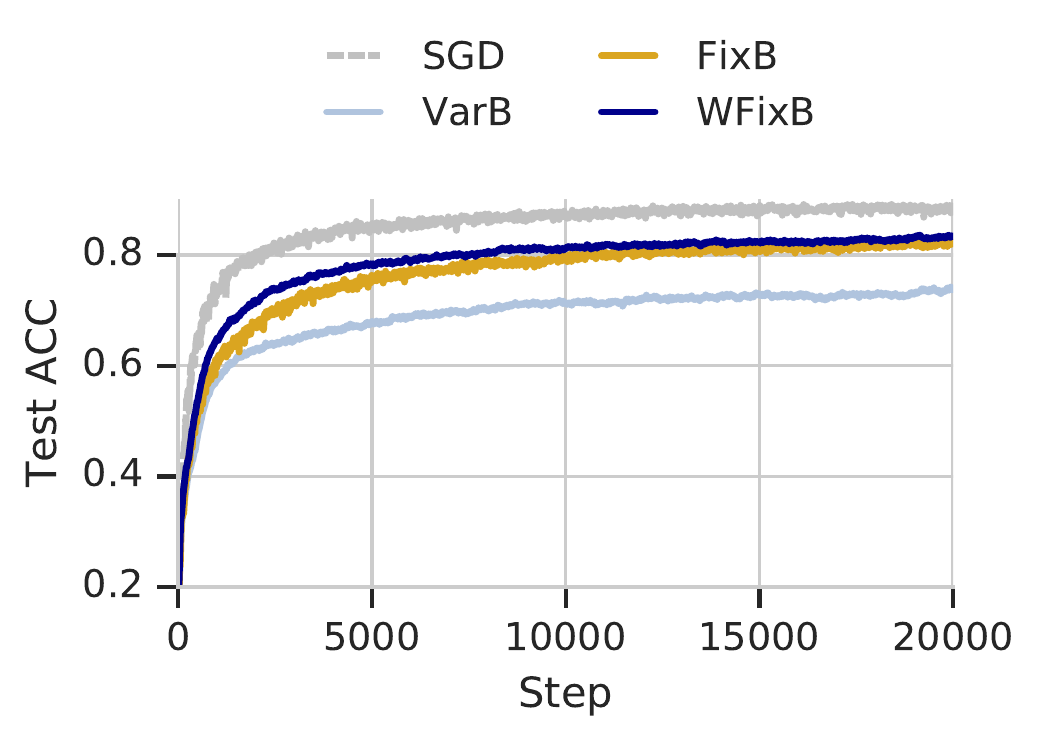}
    \end{minipage}%
    \begin{minipage}[t][][t]{.33\linewidth}        
        \includegraphics[width=1\linewidth]{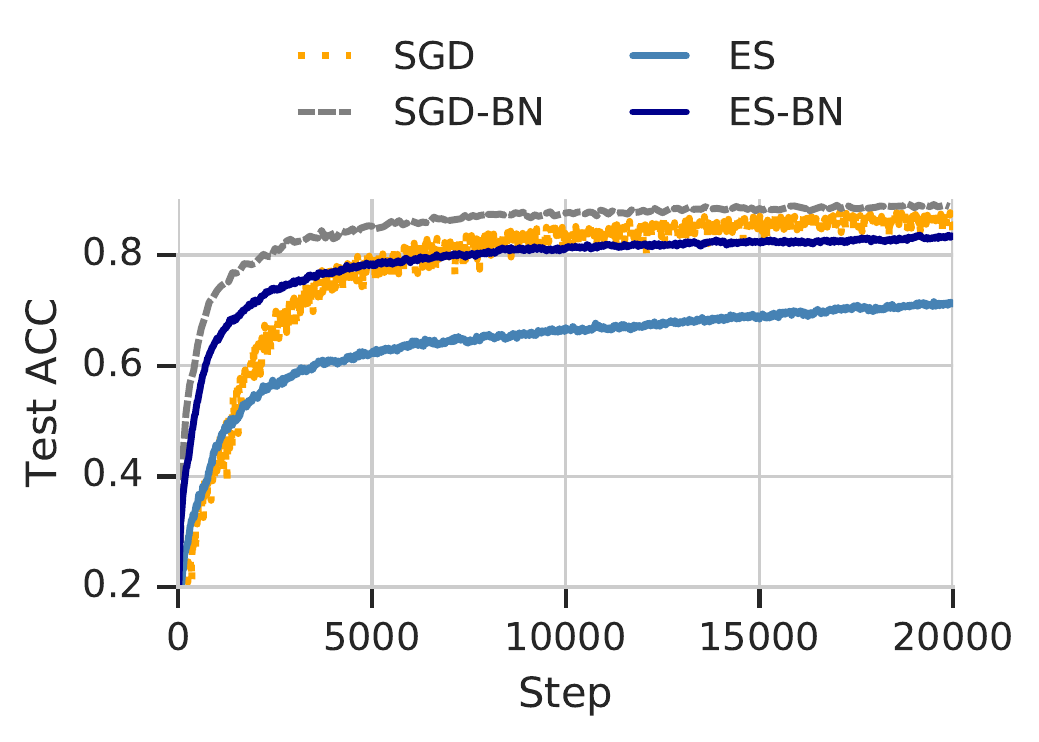}
    \end{minipage}

    \caption{
    Left - Seconds per generation of the \bigmnist model (lower is better)
    for different execution modes, trained on 110 workers with varying generation sizes $\esgsm$ (x-axis in log scale). Batches and semi-updates are computed over $\esbsm = \esgsm / 100$.
    The proposed semi-updates scheme achieves the highest training speed and allows for larger generation sizes.
    Center - Performance of \snes on \ciftinysep for different training data batch regimes. VarB computes the fitness on a variable batch per each SNES sample; FixB computes the fitness on a single batch fixed for the entire SNES generation; WFixB fixes the training batch only within each worker, and different workers use different batches.
    Right - Performance of \snes on \ciftinysep with and without Batch Normalisation. 
    }
    
    \label{fig:snes_scaling_emp}
\end{figure}

In \cref{fig:snes_scaling_emp}-left we compare the speed of different execution modes introduced in \cref{s:semi-updates}. The median time per generation is computed on a \bigmnist model on 110 workers where each worker has up to 10 CPU cores available, thus for each generation size $\esgsm$, the amount of computational resources is constant. As it can be seen, the semi-updates provide a significant speedup.

\subsection{Training on \cifar with \snes}\label{ss:snes-cifar-exp}

We have found that using \snes for training models on \cifar dataset requires a careful selection of hyper-parameters. To achieve 99\% test set accuracy on MNIST with \snes, it is typically sufficient to use a large number of parameter samples \esgst. But for \cifar,
it turned out to be challenging to achieve performance comparable to SGD.

Due to the computational complexity, we perform most of the hyper-parameter selection experiments on a relatively small model with approx. 300k parameters -- \ciftinysep, and train it for 10k steps. It uses  a parameter-efficient architecture based on separable convolutions~\citep{chollet16xception}.
This model is bigger than \umnist, but smaller than \tfmnist. 
Each experiment is run on 50 Tesla P4 GPU workers and the average processing time is 20s per generation.
For all the experiments, the results are computed over a generation size $20\,000$. Each generation is by default evaluated on a batch of 256 training images. 

\paragraph{Batch normalization}
Somewhat surprisingly, batch normalization (BN) \citep{ioffe2015batch}, is crucial for obtaining competitive results with \snes. BN is often seen as an architectural block to simplify training with SGD, but it turns out that it is even more important for \snes training, as shown in \cref{fig:snes_scaling_emp}-right. Similar effect has also been observed for RL ES models \citep{salimans2017evolution, Mania2018}.

\paragraph{Fixed versus variable batch}
For MNIST experiments, using a different batch of training examples in each sample of a generation provides better results than using the same fixed batch, as first noted in \cite{Zhang2017}. However, this does not hold for the \cifar experiments where we have consistently observed that it is important to evaluate each \snes sample within a generation on the same training batch, as shown in \cref{fig:snes_scaling_emp}-middle. We hypothesize that this could be due to the higher complexity of \cifar classification task compared to MNIST, which leads to increased variance of the fitness values when using different batches.

With semi-updates, fitness is computed individually by each worker, so it is important to fix the training data only within each worker for each semi-update. In fact, fixed batch per semi-update, WFixB, obtains slightly better performance than FixB (\cref{fig:snes_scaling_emp}-middle) due to more training data per generation. It also allows for a simpler implementation as the training data does not have to be synchronized among the workers.

\subsection{Convergence and generation size}\label{ss:es-nsamples}
Finally, we test the \snes performance versus the number of parameter samples \esgst per generation. In all experiments which we have performed with \snes, this has proven to be the most important parameter for improving performance, as observed in \cite{Zhang2017}.
For the MNIST models, we have run the training for 10k steps. Results are shown in \cref{tab:snes_nsamples}-left. For the \umnist model, it is possible to achieve a slightly better accuracy of $99\%$ at 10k steps vs $98.7\%$ in \cite{Zhang2017}.
For the \tfmnist we are able to achieve higher performance ($98.81\%$ test set acc.) than \cite{Zhang2017} mainly due to larger number of training steps, which was facilitated by the more efficient semi-updates execution model.

\begin{table}[h]
    \centering
    \caption{Test set accuracy after 10k and 20k training steps for MNIST (left) and \cifar (right) dataset respectively with SGD and \snes with various generation sizes \esgst.}
    \label{tab:snes_nsamples}
    \footnotesize
    \setlength{\tabcolsep}{2pt}
    
    \begin{tabular}{c c c}
        \begin{tabular}{c  c  c c c c c}\toprule
             & SGD &\multicolumn{5}{ c }{Generation size \esgst}  \\
             Model & Acc & 1k & 5k & 10k & 20k & 50k  \\ \midrule
             \umnist & 99.04
             & 98.57 & 98.76 & 99.13 & 99.18 & 99.16 \\ 
             \bigmnist & 99.28
             & 97.36 & 98.87 & 98.84 & 99.19 & 99.08 \\ 
             \tfmnist & 99.33
             & 96.54 & 98.46 & 98.74 & 98.60 & 98.81 \\
             \bottomrule
        \end{tabular}
        & ~ &
        \begin{tabular}{c  c   c c c c}\toprule
             & SGD &\multicolumn{4}{ c }{Generation size \esgst}  \\
             Model & Acc & 10k & 50k & 100k & 200k \\ \midrule
             \ciftinysep & 88.7
             & 80.03 & 86.32 & 88.17 & 89.48 \\ 
             \cifsep & 88.79
             & 81.48 & 87.08 & 88.47 & 89.08 \\ 
             \bottomrule
        \end{tabular}
    \end{tabular}
\end{table}

In \cref{tab:snes_nsamples}-right, we show the accuracy obtained with \snes for the \cifar models. The models use batch normalization, fixed mini-batch of 256 training images per worker (WFixB).  Similarly to the MNIST models, it is possible to reach performance comparable to SGD with a sufficient number of parameter samples. 

\paragraph{Number of training samples per evaluation}
Similarly to SGD \cite{Smith2017}, \snes performance can be improved by evaluating fitness function on more training examples (batch size), as can be seen in \cref{tab:snes_batchsize}.
We hypothesize that this is due to reduced variance of the fitness values. However, in our experiments generation size $\esgsm$ tended to have a larger effect on the final performance.

\begin{table}[h]
    \footnotesize
    \caption{Performance of \snes on \ciftinysep versus the size of the training mini-batch.}
    \label{tab:snes_batchsize}
    \centering
    \begin{tabular}{c| c c c} \toprule
         Batch Size & 128 & 256 & 512 \\ \midrule
         Val Acc & $78.61\%$ & $81.12\%$ & $83.03\%$ \\
         Gen [s] & 13.31 &  20.05 & 33.3 \\ \bottomrule
    \end{tabular}
\end{table}

\subsection{Discussion}

The empirical results show that with the right algorithmic and engineering changes, it is possible to scale up \snes and tackle tasks of higher complexity. We believe that with sufficient effort \snes can be scaled to even larger models than we have shown. Compared to standard SGD, \snes only needs to infer the model, which has a potential to enable training neural nets on inference-only or inference-optimized hardware.

\section{Hybrid \es for sparse supervised models}\label{sec:hybrid}

\begin{algorithm}[h] \footnotesize
 \KwIn{fitness $f$, diff. params. $\theta_{init}$, sparsity control $k$, temperature $\tau$, learning rates $\eta_{\logit}, \eta_{\theta}$, steps number $S$}
 $\logit \leftarrow 0$ \;
 \For{Step = 1 \dots $S$}{
    \For{i = 1 \dots n}{
      sample mask $\mathbf{m}_i \sim \mathcal{C}(\mathbf{p})$ (sampled $k$ times)\;
      $\theta_i \leftarrow \mathbf{m}_i \odot \theta$\;
      evaluate the fitness $f(\theta_i)$;~compute $\nabla_{\theta_i} f$\;
    }
    compute utilities $u_i$\;
    compute $\nabla_{\theta} f = \frac{1}{n}\sum_{i=1}^n  \nabla_{\theta_i} f$\;
    update differentiable weights $\theta = \theta + \eta_{\theta} \nabla_{\theta} f$\;
    compute $\nabla_{\logit} J = \frac{1}{\tau} \sum_{i=1}^{n} u_i \cdot (1 - \mathbf{p}) \odot \mathbf{m}_i$\;
    update mask distribution $\logit \leftarrow \logit +  \eta_{\logit} \nabla_{\logit} J$\;
 }
 \caption{\cates, a hybrid of ES and SGD}\label{alg:sgd_es}
\end{algorithm}

We have shown that it is possible to use \es for learning differentiable parameters of large supervised models. However, a key advantage of black-box optimization algorithms is that they are able to train models with non-differentiable parameters. 
In this section we show that it is indeed possible to combine conventional SGD optimization with \es for learning weight sparsity masks of sparse neural networks. We use SGD to train the differentiable weights, and \es for learning the masks at the same time. This leads to \emph{\cates}, a hybrid \es-SGD scheme which works as follows.
At each training step, a generation of mask samples is drawn from the sparsity mask distribution. Each mask sample zeroes out a subset of model weights, and the resulting sparse weights are then evaluated to obtain the mask fitness, used to compute the mask update with \es.
Simultaneously, each worker performs gradient descent w.r.t.\ its current non-zero weights, and the weight gradients are then averaged across all workers to perform a step of SGD with momentum.
The algorithm is specified in \cref{alg:sgd_es}.

This method, similarly to Drop-Connect \citep{wan2013regularization}, randomly zeroes model parameters. However, we replace the constant uniform distribution with a distribution optimized by ES.

\subsection{Sparsity mask distributions}
Mask samples $\mathbf{m}_i \in \{0,1\}^d, i = 1 \dots n$, are modelled with a multinomial distribution. 
To sample a mask, we draw repeatedly $\catnsm$ indices from a categorical distribution $\mathcal{C}(\mathbf{p})$, where $\catnsm$ controls the number of non-masked weights. 
We model the distribution probabilities $\mathbf{p}$ with a softmax function $p_j = \operatorname{exp}(\lval{j}/\tau) / \Sigma_c \operatorname{exp}(\lval{c}/\tau)$ where $\tau$ is temperature and $\logit \in \R^\esdimm$ is a vector of distribution parameters, learned with \es.
For each sample $j \sim \mathcal{C}(\mathbf{p})$, we set $m_j \leftarrow 1$, \ie we sample which model parameters are retained (not zeroed out), and the model is evaluated with $f(\mathbf{m \odot \theta})$.

We approximate the derivatives of the multinomial distribution with derivatives of the $\mathcal{C}$'s PMF $g(x = j \vert \mathbf{p}) = p_j$:
\begin{equation}
    \frac{\partial \ln g(x \vert \mathbf{p})}{\partial l_j} =
    \begin{cases}
             1 - p_j &\mbox{if } x=j    \\
            0 &\mbox{if } x\neq j 
            \end{cases}
\end{equation}
and an ES update from $\esgsm$ mask samples is approximated as: 
\begin{equation}
\nabla_{\logit} J \leftarrow \sum_{i=1}^{n} u_i \cdot \frac{1 - \mathbf{p}}{\tau} \odot \mathbf{m}_i
\end{equation} where $u_i$ is the utility of sample $i$. We do not use natural gradients.
The sparsity mask distribution parameters are updated with $\logit \leftarrow \logit + \eta_{\logit} \cdot \nabla_{\logit} J$ with learning rate $\eta_{\logit}$.

An alternative way of modelling weight sparsity is using separate Bernoulli distribution for each differentiable parameter \citep{Williams1992}.
However, it does not allow for controlling the overall sparsity of the network, as the sparsity of each weight is learned independently from the others.
Although it might be possible to control the sparsity of Bernoulli-parameterized masks using a special form of regularization as in~\citep{louizos2018learning}, in this work we opted for modelling the sparsity masks with a multinomial distribution as described above, as it allows for a direct control over the sparsity. Details about sampling without replacement are in the following section.

\subsection{Sampling from multinomial distributions}\label{ss:cat-samplers}
A mask $\mathbf{m}$, sampled from a multinomial distribution, is an outcome of $\catnsm$ draws from a categorical distribution $\mathcal{C}$.
We have observed that the implementation of the mask sampler is crucial. Not only it does affect the ability to scale to large models, but it also has a significant influence on the final performance.

The standard multinomial sampler, referred to as \cateswr, implements sampling from categorical distribution with replacement. For this method, as the number of sampler invocations $\catnsm$ increases, fewer unique indices are sampled due to the increased number of collisions, \ie $\| \mathbf{m} \|_1 < \catnsm$. 

Fixed sparsity can achieved with various methods.
As a baseline which achieves $\| \mathbf{m} \|_1 = \catnsm$, we sample  $\catnsm - \| \mathbf{m} \|_1$ additional unique non-zero indices, uniformly from the non-masked ones. This method is referred to as \catesunif. This method does not sample exactly from the original distribution but gives uniform mass to the remaining samples.

However, in our experiments sampling exactly $\catnsm$ indices without replacement tends to yield better results. Unfortunately, we have not found any computationally efficient GPU implementation of this, so instead we consider two different approximations.
The first, \catesworbatched, splits $\catnsm$ into $m$ batches. For each batch we sample $\catnsm/m$ indices using \cateswr and remove them from the distribution. This method has the advantage that for $m = \catnsm$, it converges to exact sampling without replacement, while for small $m$ it is more computationally efficient. However, unless $m = \catnsm$, it does not guarantee $\| \mathbf{m} \|_1 = \catnsm$.

In the other approximation, \catestopn, we sample $M \times \catnsm$ indices from the categorical distribution (where $M>1$ is a sufficiently large number), accumulate the sampled indices in a histogram and use its top-$\catnsm$ indices to form the mask. To break possible ties, we add small uniform noise $\mathcal{U}(0, 10^{-3})$ to the histogram\footnote{Stable sort, used in many top-N implementations, is by definition biased towards lower indices.}.

\paragraph{Fast sampler implementations}
An efficient implementation of the random sampler for $\mathcal{C}(\mathbf{p})$ is crucial for the execution speed. 
The current multinomial sampler in Tensorflow is based on Gumbel-softmax sampling \citep{maddison2017concrete, jang2017gumbel}. 
The downsides are that it requires generating $k \esdimm$ random values and the same amount of storage.  When $\esdimm$ is in the order of millions and $k$ itself is a tenth of that size (for $90\%$ sparse models), the product $\frac{\esdimm^2}{10}$ is prohibitively large, both in terms of the computation cost and memory.
A more efficient strategy is to use inverse CDF sampling. It computes the CDF of the distribution using a prefix sum \cite{blelloch1990pre, harris2007parallel} in $O(\esdimm)$ time, then generates $k$ random numbers and performs a sorted binary search into the CDF in $O(k \log \esdimm)$ with storage requirements only $O(k + \esdimm)$.
We employ CUB\footnote{\url{https://nvlabs.github.io/cub/}} to implement fast prefix sums, and use existing fast GPU RNG generators to implement a fast GPU binary search similar to the one in Thrust~\cite{bell2012thrust}. 

\subsection{Experimental results}
In this section we evaluate \cates for training sparse feed-forward and recurrent models. 
First, we deploy the proposed method on feed-forward \cifar classification models \ciftinysep and \cifbig (see \cref{ss:models}), and sparsify all layers apart from the last, fully connected layer.
We then show that \cates is also applicable to a different kind of models, 
and use it to train \wavernn~\cite{kalchbrenner2018efficient}, which is a state-of-the-art recurrent text-to-speech model.

\subsubsection{Feed-forward models}\label{ss:exps-ff}
In this section we show results for the feed-forward image classification models. 
First we investigate hyper-parameter selection, and then we compare the results against the pruning baseline \cite{NarangDSE17} which progressively masks model parameters with magnitude close to zero.
By default, in all the following experiments with \cates training, the models are trained from the start with an initial sparsity $50\%$. The ES learning rate is set to $\eta_{\logit} = 0.1$ and softmax temperature is set to $\tau = 3$ and $\esgsm=9$. We use a single \cates distribution for all weights. At test time, the mask is formed by the top-$\catnsm$ indices from $\logit$.

\paragraph{Sampling methods}

In \cref{tab:cates_samplers} we show results for different sampling methods introduced in \cref{ss:cat-samplers}. This experiment is performed on the \ciftinysep for $k = \esdimm / 2$ ($50\%$ sparsity), which is trained for 10k training steps. As can be seen, \cateswr, which does not ensure constant sparsity, reaches the worst performance. With \catesunif, where we sample additional indices from a uniform distribution, the final accuracy is considerably improved. The variants of \catesworbatched increase the accuracy even further, and the best performing method is \catestopn. We believe this is due to the fact that this method amplifies higher probabilities, and keeps a constant number of non-zero weights while still sampling model parameters with lower mask probabilities from the distribution.

\begin{table}[h]
    \centering
    \footnotesize
    \caption{Performance versus sampling method after 10k training steps, $\esgsm=200$. Dense test accuracy $87.65\%$.}
    \label{tab:cates_samplers}
    \setlength{\tabcolsep}{3pt}
    \begin{tabular}{c| c c  | c c c | c c c} \toprule
         Test Acc & \multirow{2}{*}{wR} & \multirow{2}{*}{wR+u} 
         & \multicolumn{3}{c|}{woRb, m=} & \multicolumn{3}{c}{tN, M=} \\
         $[\%]$ & & & $3$ & $4$ & $5$ & $2$ & $3$ & $5$ \\ \midrule
         \ciftinysep & $61.8$ & $74.2$ & $83.5$ & $83.4$ & $83.7$ & $85.6$ & $87.6$ & $88.6$ 
         \\ \bottomrule
    \end{tabular}
\end{table}

In all experiments of the main text, we use \catestopn approximation with $M=5$.

\paragraph{\cates and generation size}
Empirically, we have observed that the generation size has a surprisingly limited effect on \cates training, as can be seen in \cref{tab:cates_cifsep_gs}. As ConvNets are often trained with DropOut \cite{srivastava14a} or its close variant DropConnect \cite{wan2013regularization} as a form of regularization, they are known to be robust against randomly dropped weights or activations. We hypothesize that because of this robustness, finding a sparsity mask distribution for a current local optima of the differentiable weights, towards which the SGD converges, is a considerably simpler task.
Conveniently, small $\esgsm$ allows us to make use of standard pipelines for multi-GPU SGD training.

\begin{table}[h]
    \centering
    \footnotesize
    \caption{Test set accuracy after 20k training steps of \cates for 50\% sparsity with different generation size \esgst.
    }
    \label{tab:cates_cifsep_gs}
    \setlength{\tabcolsep}{4pt}
    \begin{tabular}{c | c c c c c c} \toprule
        Test Acc & \multicolumn{6}{c}{Generation Size \esgst} \\
         $[\%]$ & 2 & 5 & 10 & 50 & 100 & 200  \\ \midrule
        \ciftinysep & 87.33 & 88.13 & 88.48 & 88.54 & 88.32 & 88.72
        \\ \bottomrule
    \end{tabular}
\end{table}

\paragraph{Comparison to pruning}
In this section we compare \cates against a more conventional pruning-based method for training sparse models~\cite{NarangDSE17}.
For \cates, the update of the distribution is computed over 9 parameter samples per generation. 
We use a similar sparsity schedule for both algorithms: the models are trained for $2000$ steps with the initial sparsity ($0\%$ for pruning and $50\%$ for \cates), and then the sparsity is decreased monotonically until it reaches the final sparsity at step 50k. Overall, training is performed for 80k steps, and we use SGD with momentum, weight decay, and the learning rate follows a schedule\footnote{$0.1$ for 40k steps, then $0.01$, and $0.001$ for the last 20k steps.}.

The results of the \cates and pruning methods and the execution speed are summarized in \cref{tab:cifar_cates}-left.
Generally, \cates provides competitive results to the pruning baseline.
As \cates allows us to train sparse models from the first training step, we test different initial sparsities in \cref{tab:cifar_cates}-right, which shows that \cates is remarkably stable across initial sparsities even for models trained with only $10\%$ dense weights from scratch. In this experiment we additionally compare against a model trained with a fixed mask distribution -- ``FixMask''-- where we set the ES learning rate to zero. It shows that training of sparse over-parameterised models, such as \cifbig is possible even with a fixed sparsity mask, however it fails for models with fewer parameters where learning the mask is important.

\begin{table}[h]
    \centering
    \caption{Accuracy on the test set of the \cates trained on \cifar with constant initial sparsity $50\%$ (Left) and a constant target sparsity of $90\%$ (right). Results after 80k steps. FPS for the \cates method measured on 3 NVIDIA Tesla P100 GPU.
    }
    \label{tab:cifar_cates}
    \footnotesize
    \setlength{\tabcolsep}{2pt}
    \begin{minipage}[t]{.49\linewidth}
        \begin{tabular}{l l  c c c c  c} \toprule
            &   & \multicolumn{4}{c }{Final sparsity} &   \\
            Model & Method  & $0\%$ & $50\%$ & $80\%$ & $90\%$ & FPS \\ \midrule 
            \multirow{2}{*}{\ciftinysep} & Pruning 
            & \multirow{2}{*}{91.73} 
            & 90.98 & 88.02 & 80.62  
            & 30.6\\
            & \cates
            &
            & 89.13 & 84.98 & 79.58
            & 3.4\\
            \midrule
            
            \multirow{2}{*}{\cifsep} & Pruning 
            & \multirow{2}{*}{92.01} 
            & 90.67 & 88.37 & 81.74  
            & 25.8\\
            & \cates
            &
            & 89.01 & 86.1 & 81.94
            & 2.6\\
            \midrule
            
            \multirow{2}{*}{\cifbig} & Pruning 
            & \multirow{2}{*}{94.6} 
            & 93.32  & 92.64 & 92.24
            & 20.7\\
            & \cates
            & 
            & 93.94 & 92.87 
            & 90.83 
            & 1.32\\
            \bottomrule 
        \end{tabular}
    \end{minipage}~~%
    \begin{minipage}[t]{.49\linewidth}
        \begin{tabular}{ c  c  c c c c c} \toprule
            & & \multicolumn{5}{ c }{Initial sparsity} \\
             Model & Method & $10\%$ & $30\%$ & $50\%$ & $70\%$ & $90\%$ \\ \midrule
             \multirow{2}{*}{\ciftinysep} 
             & \cates &  80.15 & 79.46 & 79.86 & 80.23 & 79.58 \\
             & \cmask & 69.84 & 64.22 & 71.28 & 71.24 & 65.79 \\
             \midrule
             
            \multirow{2}{*}{\cifsep} 
             & \cates & 80.87 & 81.73 & 81.94 & 82.61 & 83.76\\
             & \cmask & 69.69 & 72.91 & 74.78 & 74.67 & 74.74 \\
             \midrule
             
             \multirow{2}{*}{\cifbig} 
             & \cates & 90.21 & 90.16 & 90.56 & 90.96 & 91.77 \\
             & \cmask & 90.01 & 90.01 & 90.37 & 90.59 & 90.46 \\
             \bottomrule
        \end{tabular}
    \end{minipage}
\end{table}

\subsubsection{Recurrent models}
In this section we show the results on a large sparse recurrent network, \wavernn~\cite{kalchbrenner2018efficient} for text-to-speech task. 
We trained it on a dataset of 44 hours of North American English speech recorded by a professional speaker.
The generation is conditioned on conventional linguistic features and pitch information. All compared models synthesize raw audio at 24 kHz in 16-bit format. The evaluation is carried out on a held-out test set.
We perform experiments on two models -- one with 448 (\wrnnsmall) and another with 1792 hidden state variables (\wrnnbig). As in~\cite{kalchbrenner2018efficient}, we do not sparsify the 1D convolutions at the network input, which has approximately $8M$ parameters. In total, the \wrnnsmall has $420k$ and \wrnnbig $23.7M$ masked parameters. 

For all experiments with the \cates training, models are trained with an initial sparsity $50\%$ and generation size $\esgsm = 8$ on 8 GPUs.
The sparsity decreases after 40k steps and reaches the final sparsity after 251k steps. Model parameters are trained with ADAM optimizer and a constant learning rate $2 \cdot 10^{-4}$. Otherwise, we use the same \cates hyper-parameters as in  \cref{ss:exps-ff}. 
We use a separate mask distribution per each parameter tensor which offers slightly better execution speeds. 
As noted in~\cite{kalchbrenner2018efficient}, in practical applications it might be beneficial to have the sparsity pattern in the form of contiguous blocks, so
we train the models with different sparsity block widths (width $1$ corresponds to unconstrained sparsity pattern as in the experiments above). This is implemented by using the same sparsity mask value for several contiguous weights.

\begin{table}[h]
    \centering
    \caption{NLL on the test set at 300k steps (lower is better) of the recurrent Sparse WaveRNN models. Left - Comparison of pruning and \cates on various models and block widths. Processing speed (FPS) is measured on 8 NVIDIA Tesla P100 GPU, for block width 1. Right - \cates versus FixMask trained on \wrnnbig for different initial sparsities, block width, and target sparsity of $97\%$.}
    \label{tab:res_wavernn}
    \footnotesize
    \setlength{\tabcolsep}{2pt}
    
    \begin{minipage}[t]{.49\linewidth}
        \begin{tabular}{l l | c | c | c c  c | c } \toprule
            \multirow{2}{*}{Model} & \multirow{2}{*}{Method} & SGD & \multirow{2}{*}{Spar.} &  \multicolumn{3}{c |}{Block Width}  & \multirow{2}{*}{FPS}   \\
             & & NLL &   &  1 & 2 & 16 &  \\\midrule
            \multirow{2}{*}{\wrnnsmall} & Pruning & \multirow{2}{*}{5.72} & \multirow{2}{*}{$50\%$} 
            & 5.81 & 5.78  & 5.75
            & 1.16 \\
            & \cates & & 
            & 6.02 & 5.91 & 5.80
            & 1.06  \\
            \midrule
            \multirow{2}{*}{\wrnnbig} & Pruning & \multirow{2}{*}{5.43} & \multirow{2}{*}{$97\%$}
            & 5.49 & 5.48 & 5.52
            & 0.49 \\
            & \cates & & 
            & 5.64 & 5.61 & 5.56
            & 0.45  \\
            \bottomrule 
        \end{tabular}
    \end{minipage}%
    \begin{minipage}[t]{.49\linewidth}
        \centering
        \setlength{\tabcolsep}{2pt}
        \begin{tabular}{ c  l  | c c c c} \toprule
            & & \multicolumn{3}{ c }{Initial sparsity} \\
             Model & Method & $20\%$ & $50\%$ & $80\%$ & $90\%$ \\ \midrule
             \multirow{2}{*}{\wrnnbig BW1} 
             & \cates & 5.66 & 5.64 & 5.63 & 5.63 \\
             & \cmask & 5.73 & 5.72 & 9.54 & 9.54 \\ \midrule
             \multirow{2}{*}{\wrnnbig BW16} 
             & \cates & 5.58 & 5.56 & 5.56 & 5.57  \\
             & \cmask & 5.65 & 5.65 & 9.54 & 9.54 \\ 
             \bottomrule
        \end{tabular}
    \end{minipage}%

\end{table}

The results are shown in \cref{tab:res_wavernn}. 
The computational overhead of \cates is approximately equal to the generation size, however \cates is easily parallelizable over multiple computational devices.
With each mask sample being evaluated on a single GPU, the execution speed is comparable to pruning, even though more than 57M random numbers have to be generated per training step on each GPU.
In all cases, the \cates method is competitive with the pruning baseline. In general, it performs better for larger block widths due to the reduced number of \cates parameters.
However, contrary to pruning, \cates allows us to train the sparse model from scratch -- as shown in \cref{tab:res_wavernn}-right, which opens up possibilities for using accelerated sparse training, even though we have not investigated this further here. Contrary to feed-forward models, a fixed mask  distribution (``FixMask'') does not work well for high initial sparsities.
Additionally, \cates might allow for optimizing such non-differentiable factors as execution speed.

\section{Conclusion}

In this work we have investigated the applicability of Evolution Strategies to training more complex supervised models. We have shown that using appropriate ``tricks of the trade'' it is possible to train such models to the accuracy comparable to SGD.
Additionally, we have shown that hybrid ES is a viable alternative for training sparsity masks, allowing for training sparse models from scratch in the same time as the dense models when the ES samples are parallelized across multiple computation devices.
Considering that ES is often seen as a prohibitively slow method only applicable to small problems, the significance of our results is that ES should be seriously considered as a complementary tool in the DL practitioner's toolbox, which could be useful for training non-differentiable parameters (sparsity masks and beyond) in combination with SGD. We hope that our results, albeit not the state of the art, will further reinvigorate the interest in ES and black-box methods in general.

\paragraph{Acknowledgements} We would like to thank David Choi and Jakub Sygnowski for their help developing the infrastructure used by this work.

\bibliography{literature}
\bibliographystyle{plainnat}

\nocite{buchlovsky2019tf}
\nocite{sonnetblog}

\end{document}